\documentclass{SciPost}

\usepackage[dvipsnames]{xcolor}
\hypersetup{
    colorlinks=true,
    linkcolor=BrickRed,     
    urlcolor=BrickRed,
    citecolor=BrickRed
}

\usepackage{multirow}

\usepackage[bitstream-charter]{mathdesign}
\DeclareSymbolFont{usualmathcal}{OMS}{cmsy}{m}{n}
\DeclareSymbolFontAlphabet{\mathcal}{usualmathcal}

\fancypagestyle{SPstyle}{
\fancyhf{}
\lhead{\colorbox{scipostblue}{\bf \color{white} ~SciPost Physics }}
\rhead{{\bf \color{scipostdeepblue} ~Submission }}

\fancyfoot[C]{\textbf{\thepage}}
}

\begin{document}

\pagestyle{SPstyle}

\begin{center}{\Large \textbf{\color{scipostdeepblue}{
jBOT: Semantic Jet Representation Clustering Emerges from Self-Distillation\\
}}}\end{center}

\begin{center}\textbf{
Ho Fung Tsoi\textsuperscript{} and
Dylan Rankin\textsuperscript{}
}\end{center}

\begin{center}
University of Pennsylvania, USA
\\[\baselineskip]
\small \{hftsoi,dsrankin\}@sas.upenn.edu
\end{center}

\section*{\color{scipostdeepblue}{Abstract}}
\textbf{\boldmath{%
Self-supervised learning, in the context of foundation model training, is a powerful pre-training method for learning feature representations without labels, which often capture generic underlying semantics from the data and can later be fine-tuned for downstream tasks.
In this work, we introduce jBOT, a pre-training method based on self-distillation for jet data from the CERN Large Hadron Collider, which combines local particle-level distillation with global jet-level distillation to learn jet representations that support downstream tasks such as anomaly detection and classification.
We observe that pre-training on unlabeled jets leads to emergent semantic class clustering in the representation space.
The clustering in the frozen embedding, when pre-trained on background jets only, enables anomaly detection via simple distance-based metrics, and the learned embedding can be fine-tuned for classification with improved performance compared to supervised models trained from scratch.
}}

\vspace{\baselineskip}

\noindent\textcolor{white!90!black}{%
\fbox{\parbox{0.975\linewidth}{%
\textcolor{white!40!black}{\begin{tabular}{lr}%
  \begin{minipage}{0.6\textwidth}%
    {\small Copyright attribution to authors. \newline
    This work is a submission to SciPost Physics. \newline
    License information to appear upon publication. \newline
    Publication information to appear upon publication.}
  \end{minipage} & \begin{minipage}{0.4\textwidth}
    {\small Received Date \newline Accepted Date \newline Published Date}%
  \end{minipage}
\end{tabular}}
}}
}


\vspace{10pt}
\noindent\rule{\textwidth}{1pt}
\tableofcontents
\noindent\rule{\textwidth}{1pt}
\vspace{10pt}

\section{Introduction}
\label{sec:intro}

In high-energy physics (HEP) experiments such as ATLAS~\cite{ATLAS:2008xda} and CMS~\cite{CMS:2008xjf} at the CERN Large Hadron Collider (LHC), identifying the originating particle of a jet from its substructure content (\textit{jet tagging}) is one of the primary analysis tasks for precision Standard Model measurements and new physics discoveries.
Unstable heavy particles are produced in high-energy collisions and can decay promptly in cascades until stable final states are reached and recorded by the detector.
The resulting outgoing particles are Lorentz-boosted in the direction of the original energetic particle and appear as a collection of coherent particles confined within a narrow cone from the collision point, referred to as a jet.
Jet tagging remains a challenging task because jet substructure is complex by nature, as a jet can contain $\mathcal{O}(100)$ or more nearby constituent particles and can be contaminated by background activity from other interactions.
Many machine learning techniques have been explored to improve jet tagging performance, including different architectural designs under supervised learning~\cite{Komiske:2018cqr,Qu:2019gqs,Moreno:2019bmu,Shlomi_2021,Qu:2022mxj,pmlr-v119-bogatskiy20a,Gong:2022lye,Li:2022xfc}.

In domains such as natural language modeling and computer vision, the paradigm has shifted predominantly toward first pre-training on large amounts of generic unlabeled data using self-supervised learning (SSL) to learn a representation space that encodes underlying features, and then fine-tuning on domain-specific labeled data for downstream tasks.
This two-stage approach seems to be more natural and has been shown to yield better performance than single-phase supervised learning.
A representative example is the masked language modeling (MLM) pretext task used by Bidirectional Encoder Representations from Transformers (BERT)~\cite{devlin-etal-2019-bert}, where the model is trained to predict masked words in sentences using large amounts of unlabeled text from diverse sources.
The learned representations capture the contextual semantic meaning of each word in relation to the others in a sentence and serve as foundational language knowledge, which improves performance when fine-tuned on labeled text such as sentiment classification.

The core of SSL is to \textit{learn a generic feature representation through observation without supervision}.
It aims to extract features from unlabeled data into an embedding space using self-supervised objectives such as contrastive learning~\cite{9157636,chen2020simple}, where the model is trained to be invariant to augmentations of the same example (positive pairs) while distinguishing different examples (negative pairs), or self-distillation~\cite{caron2021emerging,oquab2023dinov2,simeoni2025dinov3,zhou2021ibot,assran2023selfsupervisedlearningimagesjointembedding}, where a student network is trained to match the representations encoded by a teacher, with architectural and training designs to prevent information collapse (i.e., learning trivial solutions such as a constant vector).
This task-agnostic training encourages the model to encode features that preserve high-level semantics while being invariant to noise and low-level details, so the learned representations often exhibit meaningful properties, such as object segmentation in images or word semantics in text.
For downstream tasks such as classification, a classifier head can be attached to the learned embeddings and fine-tuned with supervision instead of training from raw inputs.
Because the embedding space already captures semantic structures from the data, fine-tuned models often outperform standalone models trained from scratch.
This has inspired recent developments of SSL in HEP~\cite{10.21468/SciPostPhys.12.6.188,Favaro:2023xdl,Katel:2024ygn,Golling:2024abg,Leigh:2024ked,Harris:2024sra,Sheldon:2024sbe,Hao:2025abk}.

In this work, we introduce jBOT, a method adapted from the self-distillation pre-training framework iBOT~\cite{zhou2021ibot} originally developed for computer vision, to learn jet representations that enable downstream tasks such as classification and anomaly detection.
Motivated by the growing attention to foundation models in the HEP field~\cite{Hallin:2025ywf,Golling:2024abg,Leigh:2024ked,Birk:2024knn,Harris:2024sra,Katel:2024ygn,Wildridge:2024yeg,Brehmer:2024yqw,Mikuni:2024qsr,Mikuni:2025tar}, we present jBOT as a viable self-distillation-based architectural option for jet data or similar physics objects at the CERN LHC.
Using a modest jet dataset as a proof-of-concept demonstration, we observe that semantic clustering of jet classes emerges in the representation space when pre-trained on unlabeled jet data via self-distillation.
The self-supervised features already exhibit class separation and enable, when pre-trained on background classes only, anomaly detection using simple metrics such as a distance-based score.
When fine-tuned on downstream classification tasks with labeled data, jBOT yields better performance than supervised models trained from scratch on the same data, especially when labeled data is limited.
The paper is structured as follows: Sec.~\ref{sec:related} discusses some recent developments in the field, Sec.~\ref{sec:method} describes the jBOT framework, Sec.~\ref{sec:experiment} presents experimental setup and results, and Sec.~\ref{sec:conclusion} summarizes the work with outlook.
Our code is available at \url{https://github.com/hftsoi/jbot}.

\section{Related Work}
\label{sec:related}

\textbf{Self-supervised visual learning.}
Early works such as MoCo~\cite{9157636} and SimCLR~\cite{chen2020simple} are based on contrastive objectives which pull positive pairs together in representation space and push apart negative pairs.
VICReg~\cite{bardes2022vicreg} does not require negative samples and prevents collapse by using regularizers to reduce redundancy in the representations.
Recent developments have focused on the self-distillation paradigm inspired by knowledge distillation~\cite{hinton2015distillingknowledgeneuralnetwork}.
DINO~\cite{caron2021emerging,oquab2023dinov2,simeoni2025dinov3} proposes a teacher-student architecture where collapse is prevented by applying stop-gradient to the teacher network whose weights are a slowly moving average of the student weights.
The objective is to match predictions in a projected feature space between positive pairs, and research shows that features extracted by self-supervised Vision Transformer (ViT)~\cite{dosovitskiy2020vit} exhibit semantic properties such as object segmentation in images that may not emerge under supervised learning.
Inspired by BERT's~\cite{devlin-etal-2019-bert} masked word prediction, iBOT~\cite{zhou2021ibot} extends the idea by masking image patches and additionally distilling the representations of the masked patches.

\textbf{SSL applications in HEP.}
Historically, machine learning in HEP has largely focused on training supervised models from scratch on labeled data~\cite{Qu:2019gqs,Qu:2022mxj}.
Recent work has started exploring SSL via task-agnostic pre-training on unlabeled data and then fine-tuning on labeled data for downstream tasks.
Contrastive methods such as SimCLR have been adapted in JetCLR~\cite{10.21468/SciPostPhys.12.6.188} and DarkCLR~\cite{Favaro:2023xdl}, which use physics symmetries to construct jet augmentations, and in RS3L~\cite{Harris:2024sra} which generates augmentations by re-simulation.
Masked particle modeling (MPM)~\cite{Golling:2024abg,Leigh:2024ked} proposes a pre-training objective based on predicting representations of masked particles.
J-JEPA~\cite{Katel:2024ygn} takes inspiration from joint-embedding predictive architectures~\cite{assran2023selfsupervisedlearningimagesjointembedding} for top tagging.
MACK~\cite{Sheldon:2024sbe}, adapted from VICReg~\cite{bardes2022vicreg}, and RINO~\cite{Hao:2025abk}, adapted from DINO~\cite{caron2021emerging}, propose using SSL to minimize performance difference when models trained on simulated labeled data are applied to real collision data, caused by mismodeling.
In parallel, recent efforts to build foundation models are emerging in the field~\cite{Hallin:2025ywf,Birk:2024knn,Wildridge:2024yeg,Brehmer:2024yqw,Mikuni:2024qsr,Mikuni:2025tar}.
These recent developments motivate exploring new applications in HEP and improving current methods, including this work.

\section{jBOT}
\label{sec:method}

\begin{figure}[!t]
    \centering
    \includegraphics[width=\textwidth]{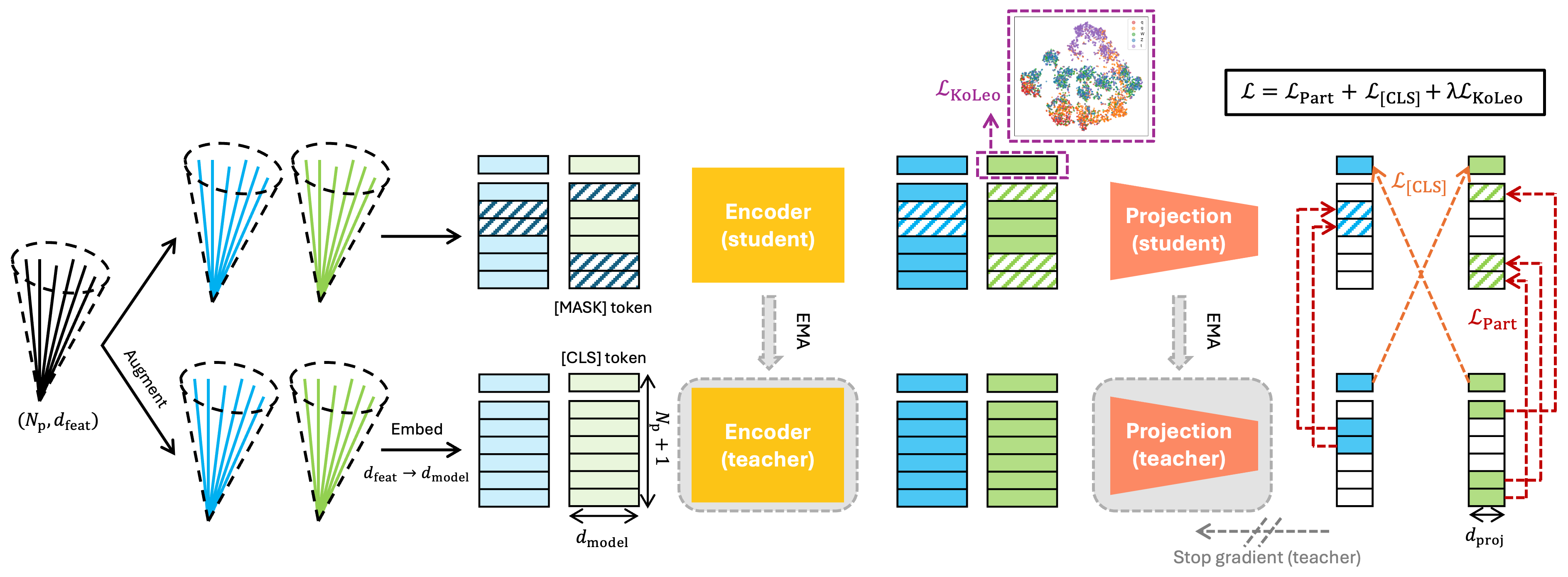}
    \caption{Schematic diagram of the jBOT pre-training method. A teacher-student architecture is used with a backbone encoder and a projection head, where stop-gradient is applied to the teacher network, whose weights are an EMA of the student network weights. Starting from an input jet, two augmented views are generated; in each view, each particle is embedded into a token space, and a \texttt{[CLS]} token is prepended. Both views are passed to the student and teacher networks. The student network processes distorted views where some of the particle tokens are masked and replaced by a learnable \texttt{[MASK]} token, while the teacher network processes the full views. Same-view and cross-view distillation losses are computed in the projection space, and the KoLeo loss is computed on the student \texttt{[CLS]} embedding from only one of the two views.}
    \label{fig:jbot}
\end{figure}

Our method largely follows iBOT~\cite{zhou2021ibot}, a self-supervised pre-training method for visual learning via self-distillation with an online tokenizer, and is adapted here to model jets with tokenized particles.
The jBOT pre-training method is schematically illustrated in Fig.~\ref{fig:jbot}, and its components are described below.

\textbf{Augmentations.}
The pre-training starts with data augmentation, which generates two views for each jet in a given batch, forming a positive pair as input to the pre-training architecture.
Following Ref.~\cite{10.21468/SciPostPhys.12.6.188}, we consider three simple augmentations: (1) uniform rotation of particles around the jet axis, (2) Gaussian smearing of particle positions, and (3) collinear splitting of particles with conserved transverse momentum ($p_{\text{T}}^{\text{initial}}=p_{\text{T}}^{\text{aug,1}}+p_{\text{T}}^{\text{aug,2}}$).

\textbf{Architecture.}
Similar to image data where fixed-sized patches are tokenized, given a jet with up to a fixed number of $N_{\text{p}}$ constituent particles, each with $d_{\text{feat}}$ features, we tokenize particles by embedding the $d_{\text{feat}}$-dimensional feature vectors into a $d_{\text{model}}$-dimensional space using a linear layer.
We additionally prepend a special learnable \texttt{[CLS]} token, which allows the network to encode global context from all other particle tokens through attention-based aggregation in the encoder, resulting in a total of $N_{\text{p}}+1$ tokens.
When jet-level features are available from the dataset and are properly reweighted to eliminate dataset priors, they can be used as conditioning inputs and embedded into the \texttt{[CLS]} token (a robust jet tagger should infer from substructure content only); otherwise the \texttt{[CLS]} token is initialized without inputs.
Each jet view therefore has an embedding shape of $(N_{\text{p}}+1, d_{\text{model}})$ and is processed by a ViT-style transformer encoder~\cite{dosovitskiy2020vit} to produce contextualized representations.
For the self-supervised objectives, a projection head is used to map the encoded tokens into a $d_{\text{proj}}$-dimensional space where distillation losses are computed.

\textbf{Self-distillation.}
The iBOT pre-training framework is formulated as knowledge distillation~\cite{hinton2015distillingknowledgeneuralnetwork} with a teacher-student architecture for learning representations via self-distillation without labels.
Both augmented views are processed by the teacher and student networks, and the student is trained to predict the teacher outputs both within the same view and across the two different views.
Unlike standard knowledge distillation, where a large teacher is pre-trained and frozen to supervise a smaller student, the teacher and student here share the same architecture and initialization and are trained jointly from scratch for the distillation objective.
To avoid collapse, an asymmetry between the teacher and student is enforced~\cite{caron2021emerging} where only the student weights $\theta_{\text{s}}$ are back-propagated, while stop-gradient is applied to the teacher whose weights $\theta_{\text{t}}$ are updated via an exponential moving average (EMA)~\cite{9157636} of the student weights by $\theta_{\text{t}}\leftarrow \tau_{\text{EMA}}\theta_{\text{t}} + (1-\tau_{\text{EMA}})\theta_{\text{s}}$, where $\tau_{\text{EMA}}\in [0,1)$ is usually set close to 1 to smoothen the teacher targets.
Subsequently in the projection space, the teacher output, $x$, is re-centered to its batch mean as $x\leftarrow x-c$ with the center updated via EMA using $c\leftarrow \tau_{\text{c}} c + (1-\tau_{\text{c}})\bar{x}$, where $\tau_{\text{c}}$ is the centering momentum and $\bar{x}$ is the batch mean.
Finally, the outputs are mapped by a temperature-scaled softmax into $d_{\text{proj}}$-dimensional distributions, so the teacher and student outputs are matched via, e.g., cross-entropy.
There are local same-view and global cross-view distillations that are described in the following.

\textbf{Particle-level objective (same-view distillation).}
In iBOT, each of the two image views passed to the student network has a fraction of patches masked by replacing the corresponding patch tokens with a learnable \texttt{[MASK]} token, so the student processes distorted views, while the teacher processes the complete views.
For jets, similar to MPM~\cite{Golling:2024abg,Leigh:2024ked}, masking is performed on a per-particle basis.
Unlike images, where a grid is drawn and the image is divided into equally sized non-overlapping patches, particles within a jet can have widely varying positions and momenta.
Therefore, instead of treating all particles equally by randomly masking a fixed number of particles, which can result in masking mostly the most energetic particles or mostly the least energetic particles, we use a simple momentum-aware scheme that masks particles such that the cumulative transverse momentum of the masked particles reaches a target ratio.
For example, a 30\% target masking ratio corresponds to randomly selecting a subset of particles to be masked such that approximately 30\% of the jet transverse momentum is masked.
For an input view $u$, the student network outputs $N_{\text{p}}+1$ vectors in the projection space, $P_{\text{s}}^{\text{part},i=1,..,N_{\text{p}}}(u)\in [0,1]^{d_{\text{proj}}}$ and $P_{\text{s}}^{\texttt{[CLS]}}(u)\in [0,1]^{d_{\text{proj}}}$, and similarly for the teacher network outputs, denoted by $P_{\text{t}}$.
For an augmented pair $(u,v)$, the student is trained to predict the teacher outputs for the masked particle tokens within the same view using a cross-entropy loss:
\begin{equation}
    \label{eq:loss-patch}
    \begin{aligned}
    &\ell_{\text{Part}}(u)=\frac{1}{M(u)}\sum_{i=1}^{N_{\text{p}}}m_i(u)\bigg(-P_{\text{t}}^{\text{part},i}(u)^{\text{T}}\cdot\log P_{\text{s}}^{\text{part},i}(u)\bigg), \\
    &\mathcal{L}_{\text{Part}}=\frac{1}{2}\big(\ell_{\text{Part}}(u)+\ell_{\text{Part}}(v)\big),
    \end{aligned}
\end{equation}
where $m_{i}(u)\in\{0,1\}$ indicates if the $i$-th particle is masked ($m_{i}=1$) or not ($m_{i}=0$), and $M(u)=\sum_{i=1}^{N_{\text{p}}}m_i(u)$ is the number of masked particles in view $u$.

\textbf{Jet-level objective (cross-view distillation).}
Complementary to the same-view distillation, which encourages the model to encode particle-level structure in the learned representations, the cross-view objective distills global representations by matching the teacher \texttt{[CLS]} output from view $u$ to student \texttt{[CLS]} output from view $v$, and vice versa:
\begin{align}
    \label{eq:loss-cls}
    \begin{aligned}
    &\ell_{\texttt{[CLS]}}(u,v)=-P_{\text{t}}^{\texttt{[CLS]}}(u)^{\text{T}}\cdot \log P_{\text{s}}^{\texttt{[CLS]}}(v),\\
    &\mathcal{L}_{\texttt{[CLS]}}=\frac{1}{2}\big(\ell_{\texttt{[CLS]}}(u,v)+\ell_{\texttt{[CLS]}}(v,u)\big).
    \end{aligned}
\end{align}

\textbf{Feature space diversification.}
In addition, similar to DINOv2~\cite{oquab2023dinov2}, we add the KoLeo regularizer~\cite{sablayrolles2018spreading} to encourage a diverse spread of different examples in the embedding space within a batch:
\begin{equation}
    \label{eq:loss-koleo}
    \mathcal{L}_{\text{KoLeo}} = -\frac{1}{B}\sum_{i=1}^{B}\log\big(\min_{j\neq i} \lVert x_j - x_i \rVert\big)
\end{equation}
where $x_i$ is a vector in the embedding space after $\ell_2$-normalization and the sum runs over a batch of size $B$.
To simplify the computation, we apply the regularizer to the student \texttt{[CLS]} from only one of the two views.

\begin{figure}[!t]
    \centering
    \includegraphics[width=0.7\textwidth]{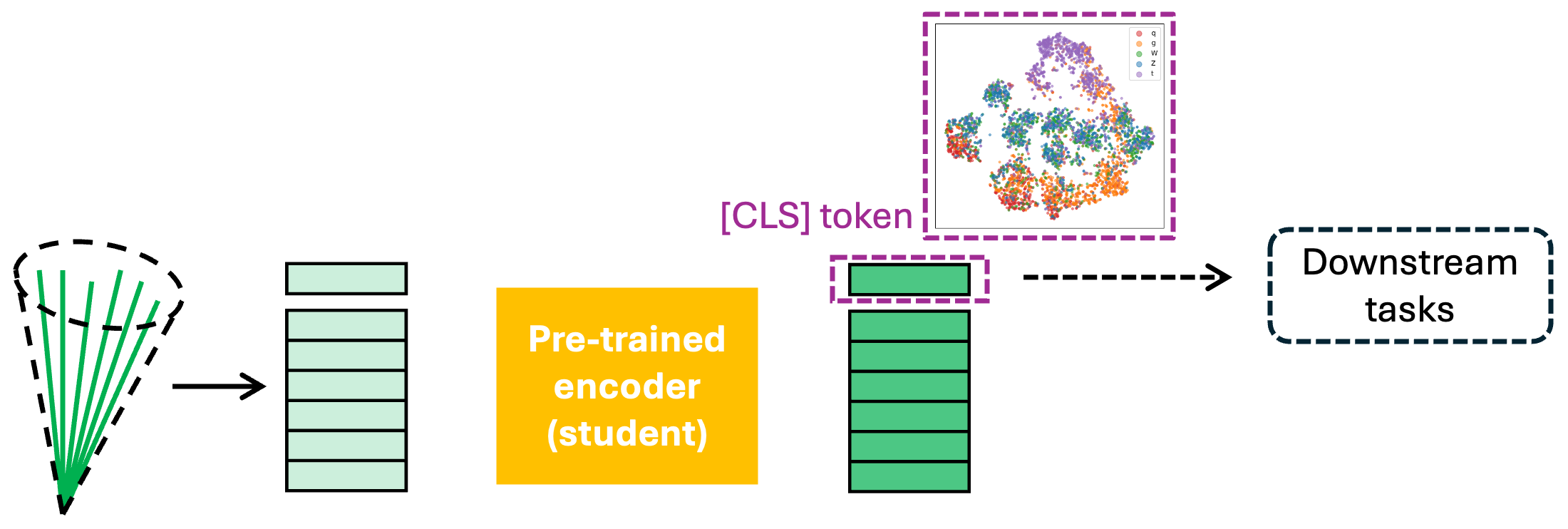}
    \caption{Downstream tasks are performed using the \texttt{[CLS]} embedding from the pre-trained student encoder.}
    \label{fig:jbot-downstream}
\end{figure}

To sum up, the pre-training loss is given by:
\begin{equation}
    \label{eq:loss-total}
    \mathcal{L}=\mathcal{L}_{\text{Part}}+\mathcal{L}_{\text{\texttt{[CLS]}}}+\lambda \mathcal{L}_{\text{KoLeo}}
\end{equation}
where $\lambda>0$ scales the KoLeo regularization.
After pre-training, the projection head is removed, and test jets without augmentations and masking are processed by the student encoder to produce self-supervised features for downstream tasks, as illustrated in Fig.~\ref{fig:jbot-downstream}.
For instance, one can attach a classifier head taking as input the encoded \texttt{[CLS]} token and fine-tune both the encoder and classifier for supervised classification, or directly probe the frozen representation for anomaly detection.

\section{Experiment}
\label{sec:experiment}

\subsection{Dataset}
\label{sec:exp-dataset}
We train and evaluate on the JetNet dataset~\cite{Kansal:2021cqp,kansal_2022_6975118}.
The dataset consists of 880k simulated jets with transverse momentum ($p_{\text{T}}$) around 1 TeV, originating from light quarks (q), gluons (g), W bosons, Z bosons, and top quarks (t) produced in proton-proton collisions at a center-of-mass energy of 13 TeV.
Jet clustering is performed using the anti-$k_{\text{T}}$ algorithm~\cite{Cacciari:2008gp} with a distance parameter of 0.8.
Each jet stores up to 30 highest-$p_{\text{T}}$ constituent particles, each with four features $(\eta_{\text{rel}}, \phi_{\text{rel}}, p_{\text{T,rel}}, \text{valid})$, where $\eta_{\text{rel}}=\eta-\eta_{\text{jet}}$ and $\phi_{\text{rel}}=\phi-\phi_{\text{jet}}$ are the pseudorapidity and azimuthal angle measured from the jet axis, $p_{\text{T,rel}}=p_{\text{T}}/p_{\text{T,jet}}$ is the $p_{\text{T}}$ fraction relative to the jet, and ``valid'' is a boolean indicating whether the particle is padded when the jet contains fewer than 30 particles.
The dataset also contains jet-level kinematic features such as jet $p_{\text{T}}$, $\eta$, and $\phi$, but we do not consider them in the models, since their distributions differ between classes and these dataset priors, without proper re-weighting, may introduce bias into jet classification, which should be based on jet substructure information only.
In spite of this, the jBOT method is capable of handling these global features.
We also note that the dataset is modest in size in the context of foundation model training, so the downstream results should be interpreted as a proof-of-concept demonstration that self-distillation on jets can yield semantic and transferable representations.

\subsection{Implementation}
\label{sec:exp-implementation}

\begin{figure}[!t]
    \centering
    \includegraphics[width=\textwidth]{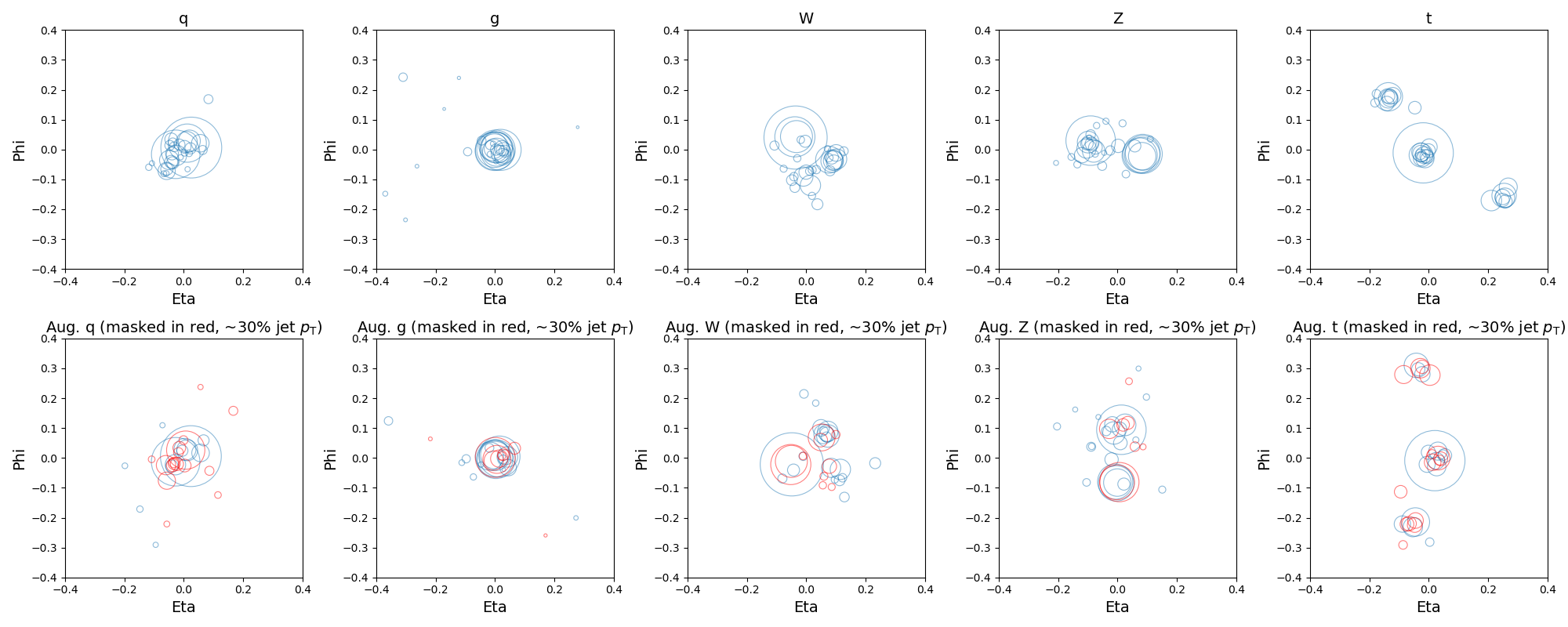}
    \caption{One example jet per class, where each circle represents a particle: the circle center is at the particle location and the radius is proportional to its $p_{\text{T}}$. Upper row: input jets. Lower row: augmented jets with masking shown in red (e.g., $\sim$30\% of the jet $p_{\text{T}}$).}
    \label{fig:jets}
\end{figure}

For the augmentations, the rotation angle is sampled uniformly from $-\pi$ to $\pi$ per jet; following Ref.~\cite{10.21468/SciPostPhys.12.6.188}, each particle's $\eta$ and $\phi$ are smeared independently by a Gaussian with a variance of $\Lambda_{\text{QCD}}/p_{\text{T}}$, where $\Lambda_{\text{QCD}}=100$ MeV is the QCD scale; and collinear splitting is applied to jets with fewer than 30 valid particles.
Masking is implemented by accumulating particles from a randomly reshuffled list until the cumulative $p_{\text{T}}$ crosses the masking target; the subset whose cumulative $p_{\text{T}}$ is closest to the target is masked, which avoids overshooting or undershooting the target masking ratio.
Fig.~\ref{fig:jets} shows example augmented jets with masking applied.

The model and pre-training hyperparameters are summarized in Tab.~\ref{tab:config}; note that these parameters are not rigorously optimized but are reasonably chosen.
We use a ViT-style transformer~\cite{dosovitskiy2020vit} as the backbone encoder, and consider two model sizes: small (jBOT-S) and base (jBOT-B).
The projection head is a multilayer perceptron (MLP), and the weights are shared for projecting both the \texttt{[CLS]} and particle tokens.
Dropout~\cite{JMLR:v15:srivastava14a} with a rate of 20\% is used in the transformer blocks.
All hidden layers use Gaussian error linear unit (GELU)~\cite{hendrycks2016gelu} activation.
The model is implemented using Tensorflow~\cite{tensorflow2015-whitepaper} and Keras~\cite{chollet2015keras}, and optimized using the AdamW optimizer~\cite{loshchilov2018decoupled}.

\begin{table*}[!t]
\caption{Model and pre-training hyperparameters.}
\label{tab:config}
\centering
\resizebox{\textwidth}{!}{
\begin{tabular}{l|l} \hline

    \multicolumn{2}{l}{\textit{Model hyperparameters}} \\ \hline
    Embedding dimension $d_{\text{model}}$ & 32 (small), 64 (base) \\
    \# of transformer blocks in encoder  &  2 (small), 4 (base)\\
    \# of self-attention heads & 4 (small), 6 (base) \\
    Feedforward layer dim. in transformer block & $4d_{\text{model}}$ \\
    Projection dimension $d_{\text{proj}}$ & $d_{\text{model}}/2$ \\
    Feedforward layer dim. in projection head  & Two hidden layers: $8d_{\text{proj}}$, $d_{\text{proj}}$\\
    Feedforward layer dim. in classifier head & Two hidden layers: $2d_\text{model}$, $d_\text{model}$ \\ \hline

    \multicolumn{2}{l}{\textit{Pre-training hyperparameters}} \\ \hline
    Masking ratio ($p_{\text{T}}$) & Uniformly sampled from 0 to 50\% per view \\
    EMA momentum $\tau_{\text{EMA}}$ & Cosine schedule from 0.996 to 1 \cite{NEURIPS2020_f3ada80d}\\
    Centering momentum $\tau_{\text{c}}$ & 0.9 \\
    Softmax temperature & 0.04 (teacher), 0.1 (student) \\
    KoLeo scale $\lambda$ & 0.01 \\
    Learning rate & $5\times10^{-4}\times (\text{batch size}/256)$, linear warm up for first 10 epochs\\
    Batch size & 1024 \\
    Weight decay & $10^{-4}$ \\ \hline
\end{tabular}%
}
\end{table*}

\subsection{Pre-training}
\label{sec:exp-pretrained}

We pre-train on three different training sets containing (1) all five jet classes, (2) the q, g, and t classes, and (3) the q and g classes, which are used downstream to probe five-class classification, top tagging, and anomaly detection, respectively.
The training/validation/test split is around 80/10/10\%, with balanced classes.
We pre-train for 100 epochs on a single Nvidia A100 GPU, which takes, for example, around 11 min/epoch and in total 18 hours for the base model on the five-class training set with 700k examples.

\begin{figure}[!t]
    \centering
    \includegraphics[width=0.7\textwidth]{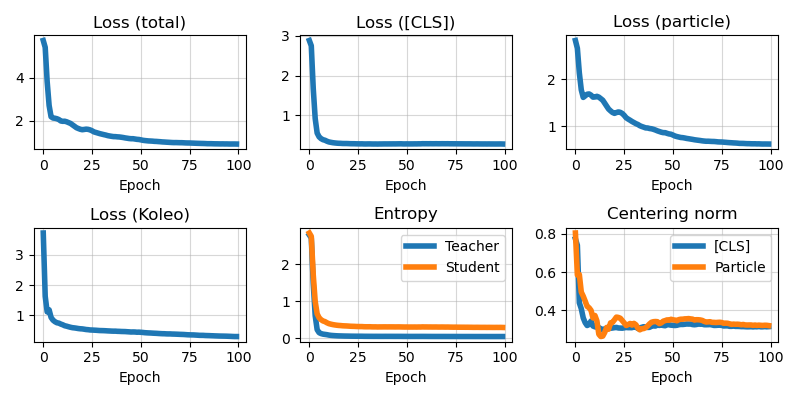}
    \caption{Learning curves for jBOT-B when pre-trained on five-class data. Upper left to right: total loss, $\mathcal{L}_{\texttt{[CLS]}}$, and $\mathcal{L}_{\text{Part}}$. Lower left to right: $\mathcal{L}_{\text{KoLeo}}$, entropy of the \texttt{[CLS]}, and centering norm in the teacher.}
    \label{fig:training_curve}
\end{figure}

The learning curves are shown in Fig.~\ref{fig:training_curve}, where losses, entropy of the \texttt{[CLS]} softmax outputs ($-P^{\texttt{[CLS]}\text{T}}\cdot\log P^{\texttt{[CLS]}}$), and the norm of the teacher centering vectors are monitored.
The losses decrease steadily and plateau.
Collapse is prevented by the balance between sharpening and centering~\cite{caron2021emerging}.
Due to the sharper teacher softmax temperature, the teacher entropy is slightly below the student entropy, and both stabilize at values well below the bound $\log (d_{\text{proj}})=\log(32)\approx 3.47$, indicating no collapse to a uniform output.
While the centering encourages a uniform output, the entropies stabilize at values above zero, indicating no collapse to a single-logit output.

\begin{figure}[!t]
    \centering
    \includegraphics[width=0.73\textwidth]{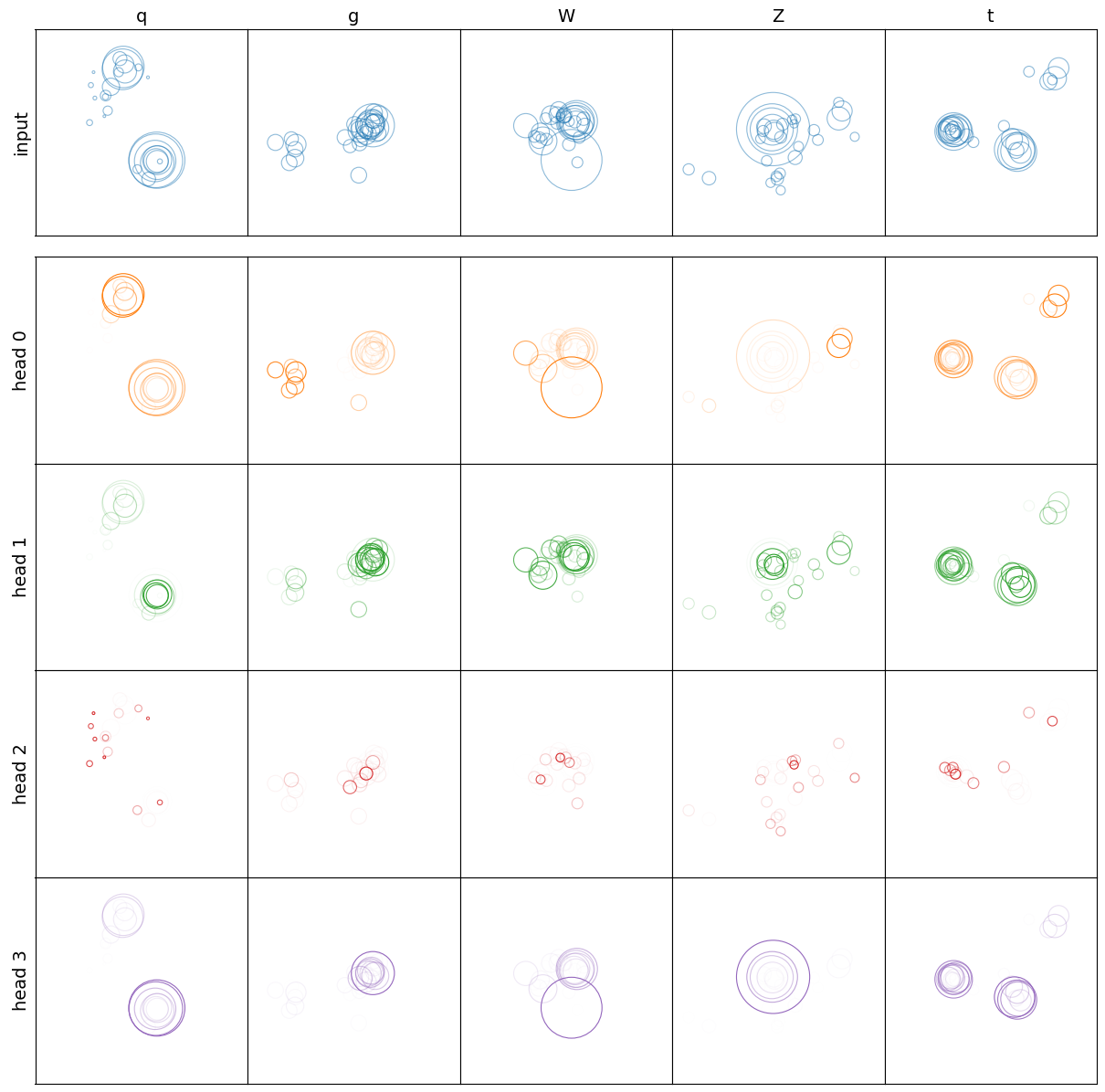}
    \caption{Attention weights from the last transformer block in jBOT-S pre-trained on all five classes, obtained using the \texttt{[CLS]} token as the query attending to the particle tokens. Top row: one example input jet per class (each circle represents a particle: the circle center is at the particle location, the radius is proportional to its $p_{\text{T}}$, and the edge alpha is uniform across all particles here). Other rows: attention weights per head for the same input jets, shown with the same drawing style as the input jets, but with the attention weight represented by the edge alpha (higher edge alpha indicates larger attention weight).}
    \label{fig:attention}
\end{figure}

Fig.~\ref{fig:attention} shows how the pre-trained encoder weights particles across attention heads by visualizing the attention weights from the \texttt{[CLS]} token as the query attending to the particle tokens in the last transformer layer.
Different heads attend to different groups of particles within a jet, say some focus on soft particles only and some on the hardest local particles that likely correspond to the prong structure, indicating that the model learns high-level features and correlations within the jet substructure.

\begin{figure}[!t]
    \centering
    \includegraphics[width=1\textwidth]{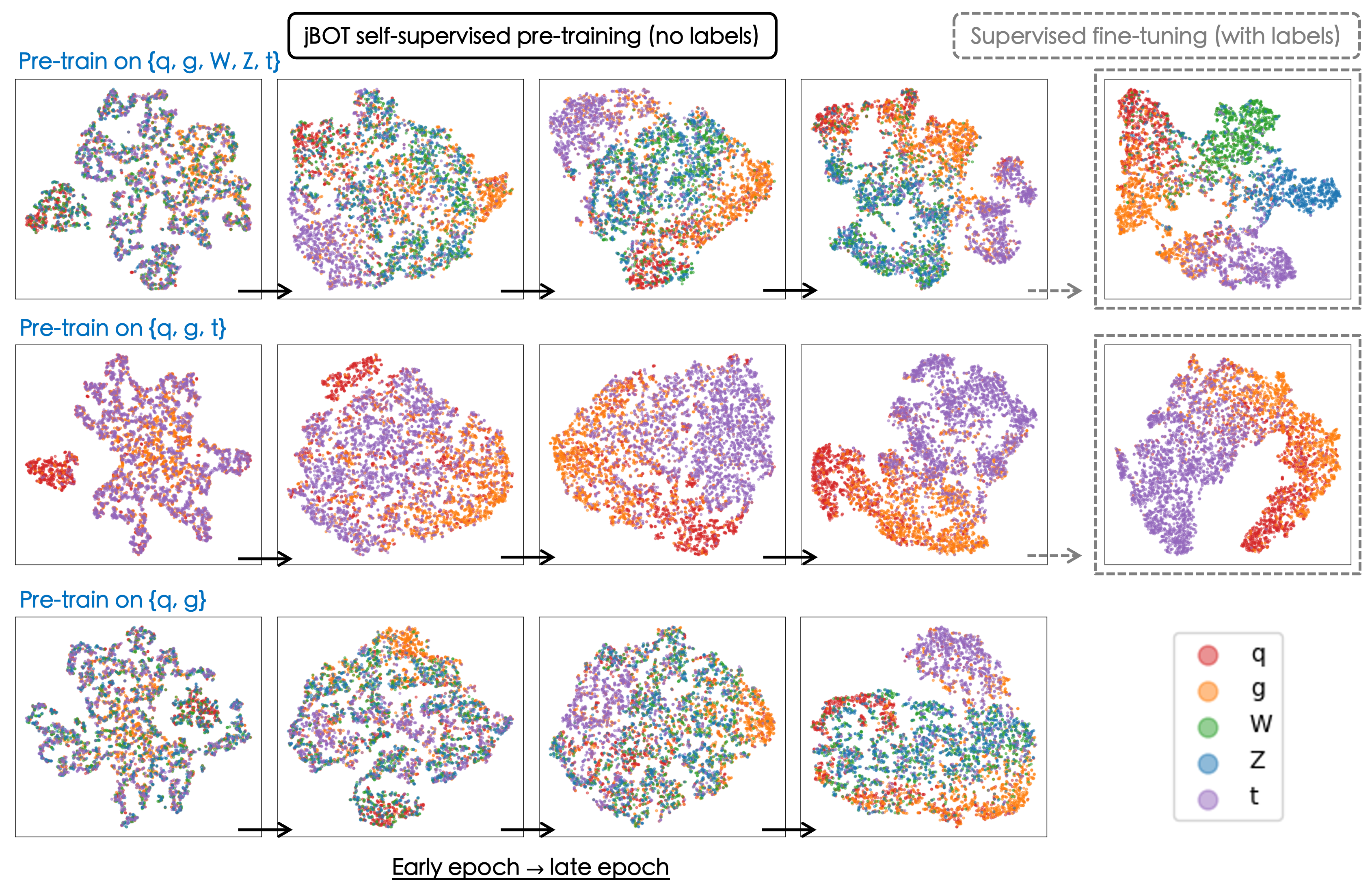}
    \caption{Example evolution of 2D t-SNE projections of the [CLS] token during pre-training without labels (and after fine-tuning with labels). Top row: pre-training on all five classes and then fine-tuning. Middle row: pre-training on the t, q, and g classes and then fine-tuning. Bottom row: pre-training on the q and g classes.}
    \label{fig:tsne}
\end{figure}

Fig.~\ref{fig:tsne} shows the evolution of the \texttt{[CLS]} embedding during pre-training using 2D t-SNE projections~\cite{JMLR:v9:vandermaaten08a}, which preserve the underlying local neighborhood structure of the embedding.
Starting from a random initialization of model parameters, where the embedding is random, the projections across epochs show the progressive formation of compact clusters separating different classes, demonstrating that the self-distillation objective yields discriminative structure even though no class labels are used during pre-training.

\subsection{Downstream classification}
\label{sec:exp-class}

We first evaluate the pre-trained encoder by probing classification on the frozen features from the \texttt{[CLS]} token.
We probe five-class classification using the student encoder pre-trained on all five classes, and top tagging using the student encoder pre-trained on the t, q, and g classes.
Following standard evaluation protocols~\cite{caron2021emerging,zhou2021ibot}, we fit a $k$-nearest-neighbor ($k$-NN) classifier with $k=30$ and a linear classifier on the frozen features.
The performance is shown in Tab.~\ref{tab:5class} (five-class) and Tab.~\ref{tab:top} (top tagging).
For example, both $k$-NN and linear probe achieve accuracies of around 70\% in the five-class set and 87\% in the top tagging set.

We then construct a classifier by attaching an MLP head that takes as input the \texttt{[CLS]} token from the pre-trained encoder and perform supervised fine-tuning on the labeled training set.
To preserve underlying semantic structures learned from pre-training and avoid degenerating into a supervised classifier trained from scratch, we use layer-wise learning rate decay (LLRD)~\cite{bao2022beit,dong2022ftclip} for the fine-tuning: the classifier head receives the largest learning rate, and the learning rate decays by a multiplicative factor in each of the earlier blocks, so that the earliest layers in the encoder receive the smallest learning rate.
We fine-tune separately using only 10\% and the full training set, scanning decay factors $\{0.6,0.65,0.7,0.75,0.8\}$ and base learning rates $\{4\times 10^{-5}, 8\times 10^{-5}, 2\times 10^{-4}, 4\times 10^{-4}, 8\times 10^{-4}, 2\times 10^{-3}\}$, with a batch size of 1024 for 100 epochs, and then select the model with the best performance.
The supervised model is trained with the same batch size, number of epochs, and the learning-rate tuning.

\begin{figure}[!t]
    \centering
    \includegraphics[width=0.48\textwidth]{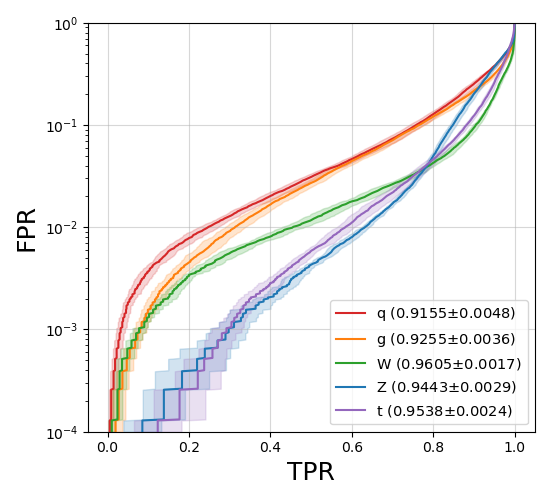}
    \includegraphics[width=0.48\textwidth]{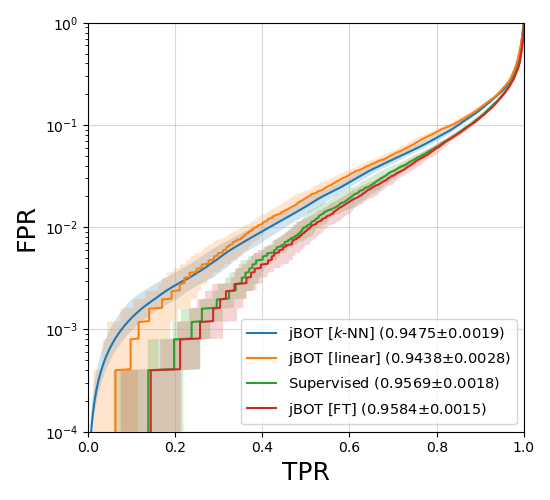}
    \caption{ROC curves for classification. Left: jBOT-B (fine-tuned) on five-class classification. Right: jBOT-B (using frozen features and fine-tuned) on top tagging.}
    \label{fig:roc-ft}
\end{figure}

The results are shown in Tab.~\ref{tab:5class} (five-class) and Tab.~\ref{tab:top} (top tagging), and ROC curves are shown in Fig.~\ref{fig:roc-ft}.
All fine-tuned models perform better than or match supervised models trained from scratch on the same labeled dataset sizes, with the largest gains seen when the fine-tuning dataset is small (e.g., 10\%), because the self-supervised models are fine-tuned from a feature representation that already encodes meaningful information.
We note that models fine-tuned on the 10\% labeled data have already been exposed to the full training set during the unlabeled pre-training stage, so the comparison should be interpreted as a probe of label efficiency in practical scenarios where unlabeled data is abundant while labeled data is scarce.
A controlled comparison with disjoint pre-training and fine-tuning datasets is to be done in future work that scales up to a larger dataset.

\begin{table*}[!t]
\caption{Five-class classification performance (overall accuracy and per-class AUC) comparing jBOT with supervised models. Note that all models use particle features only and ignore jet-level features.}
\label{tab:5class}
\centering
\resizebox{\textwidth}{!}{
\begin{tabular}{lcccccc} \hline
    \multirow{2}{*}{Model} & \multirow{2}{*}{Acc. [\%]} & \multicolumn{5}{c}{AUC} \\
    & & q & g & W & Z & t \\ \hline
    \multicolumn{2}{l}{\textit{Frozen embedding (no labels)}}  &  &  &  &  &  \\
    
    $k$-NN (jBOT-S) & 71.02 $\pm$ 0.58 & 0.8869 $\pm$ 0.0058 & 0.8943 $\pm$ 0.0040 & 0.9272 $\pm$ 0.0021 & 0.9070 $\pm$ 0.0039 & 0.9209 $\pm$ 0.0052 \\
    Linear (jBOT-S) & 67.43 $\pm$ 0.56 & 0.8702 $\pm$ 0.0049 & 0.8796 $\pm$ 0.0042 & 0.8904 $\pm$ 0.0017 & 0.8820 $\pm$ 0.0037 & 0.9166 $\pm$ 0.0044 \\ \hline
    
    $k$-NN (jBOT-B) & 70.53 $\pm$ 0.57 & 0.8859 $\pm$ 0.0049 & 0.8883 $\pm$ 0.0051 & 0.9209 $\pm$ 0.0029 & 0.9057 $\pm$ 0.0040 & 0.9218 $\pm$ 0.0044 \\
    Linear (jBOT-B) & 69.42 $\pm$ 0.46 & 0.8767 $\pm$ 0.0058 & 0.8820 $\pm$ 0.0051 & 0.9107 $\pm$ 0.0023 & 0.9019 $\pm$ 0.0034 & 0.9203 $\pm$ 0.0044 \\ \hline \hline
    
    \multicolumn{2}{l}{\textit{Fine-tuning (with labels)}}  &  &  &  &  &  \\

    \textcolor{gray}{Sup.-S (10\%)}  & \textcolor{gray}{72.51 $\pm$ 0.52} & \textcolor{gray}{0.8908 $\pm$ 0.0057} & \textcolor{gray}{0.8966 $\pm$ 0.0039} & \textcolor{gray}{0.9463 $\pm$ 0.0020} & \textcolor{gray}{0.9290 $\pm$ 0.0035} & \textcolor{gray}{0.9335 $\pm$ 0.0040} \\
    jBOT-S (10\%)                    & 74.25 $\pm$ 0.56 & 0.9004 $\pm$ 0.0055 & 0.9061 $\pm$ 0.0035 & 0.9541 $\pm$ 0.0022 & 0.9359 $\pm$ 0.0031 & 0.9427 $\pm$ 0.0034 \\

    \textcolor{gray}{Sup.-S (100\%)} & \textcolor{gray}{74.31 $\pm$ 0.56} & \textcolor{gray}{0.8987 $\pm$ 0.0056} & \textcolor{gray}{0.9070 $\pm$ 0.0037} & \textcolor{gray}{0.9535 $\pm$ 0.0023} & \textcolor{gray}{0.9359 $\pm$ 0.0032} & \textcolor{gray}{0.9427 $\pm$ 0.0033} \\
    jBOT-S (100\%)                   & 75.26 $\pm$ 0.64 & 0.9074 $\pm$ 0.0051 & 0.9177 $\pm$ 0.0035 & 0.9573 $\pm$ 0.0022 & 0.9387 $\pm$ 0.0030 & 0.9477 $\pm$ 0.0030 \\ \hline
    
    \textcolor{gray}{Sup.-B (10\%)}  & \textcolor{gray}{73.79 $\pm$ 0.56} & \textcolor{gray}{0.9007 $\pm$ 0.0051} & \textcolor{gray}{0.9070 $\pm$ 0.0037} & \textcolor{gray}{0.9517 $\pm$ 0.0023} & \textcolor{gray}{0.9333 $\pm$ 0.0036} & \textcolor{gray}{0.9450 $\pm$ 0.0038} \\
    jBOT-B (10\%)                    & 74.99 $\pm$ 0.53 & 0.9078 $\pm$ 0.0048 & 0.9174 $\pm$ 0.0038 & 0.9557 $\pm$ 0.0021 & 0.9372 $\pm$ 0.0027 & 0.9479 $\pm$ 0.0032 \\

    \textcolor{gray}{Sup.-B (100\%)} & \textcolor{gray}{76.04 $\pm$ 0.57} & \textcolor{gray}{0.9135 $\pm$ 0.0052} & \textcolor{gray}{0.9231 $\pm$ 0.0036} & \textcolor{gray}{0.9596 $\pm$ 0.0018} & \textcolor{gray}{0.9430 $\pm$ 0.0028} & \textcolor{gray}{0.9518 $\pm$ 0.0028} \\
    jBOT-B (100\%)                   & \textbf{76.43 $\pm$ 0.61} & \textbf{0.9155 $\pm$ 0.0048} & \textbf{0.9255 $\pm$ 0.0036} & \textbf{0.9605 $\pm$ 0.0017} & \textbf{0.9443 $\pm$ 0.0029} & \textbf{0.9538 $\pm$ 0.0024} \\ \hline
\end{tabular}%
}
\end{table*}

\begin{table*}[!t]
\caption{Top tagging performance (accuracy, AUC, signal efficiency $\epsilon_{\text{s}}(X)$ at fixed background efficiency $\epsilon_{\text{b}}=X$, and background rejection $R_{Y\%}=1/\epsilon_{\text{b}}$ at fixed $\epsilon_{\text{s}}=Y\%$) comparing jBOT with supervised models. Note that all models use particle features only and ignore jet-level features.}
\label{tab:top}
\centering
\resizebox{\textwidth}{!}{
\begin{tabular}{lcccccc} \hline
    Model & Acc. [\%] & AUC & $\epsilon_{\text{s}}(10^{-1})$ & $\epsilon_{\text{s}}(10^{-2})$ & $R_{30\%}$ & $R_{50\%}$ \\ \hline
    \multicolumn{2}{l}{\textit{Frozen embedding (no labels)}}  &  &  &  & &  \\
    
    $k$-NN (jBOT-S) & 87.77 $\pm$ 0.37 & 0.9447 $\pm$ 0.0023 & 0.8352 $\pm$ 0.0125 & 0.3337 $\pm$ 0.0572 & 133.0 & 42.4  \\
    Linear (jBOT-S) & 87.09 $\pm$ 0.39 & 0.9355 $\pm$ 0.0020 & 0.8115 $\pm$ 0.0138 & 0.2408 $\pm$ 0.0327 & 74.9 & 31.8  \\ \hline
    
    $k$-NN (jBOT-B) & 87.93 $\pm$ 0.35 & 0.9475 $\pm$ 0.0019 & 0.8402 $\pm$ 0.0066 & 0.3494 $\pm$ 0.0699 & 100.8 & 60.4  \\
    Linear (jBOT-B) & 87.65 $\pm$ 0.41 & 0.9438 $\pm$ 0.0028 & 0.8344 $\pm$ 0.0116 & 0.3892 $\pm$ 0.0315 & 178.6 & 51.8  \\ \hline \hline
    
    \multicolumn{2}{l}{\textit{Fine-tuning (with labels)}}  &  &  &  &  &   \\

    \textcolor{gray}{Sup.-S (10\%)}  & \textcolor{gray}{87.23 $\pm$ 0.40} & \textcolor{gray}{0.9368 $\pm$ 0.0028} & \textcolor{gray}{0.8172 $\pm$ 0.0158} & \textcolor{gray}{0.2166 $\pm$ 0.0394} & \textcolor{gray}{64.3} & \textcolor{gray}{30.9}   \\
    jBOT-S (10\%)                    & 88.07 $\pm$ 0.47 & 0.9499 $\pm$ 0.0019 & 0.8559 $\pm$ 0.0106 & 0.4306 $\pm$ 0.0317 & 271.7 & 61.0   \\

    \textcolor{gray}{Sup.-S (100\%)} & \textcolor{gray}{88.52 $\pm$ 0.47} & \textcolor{gray}{0.9524 $\pm$ 0.0016} & \textcolor{gray}{0.8659 $\pm$ 0.0080} & \textcolor{gray}{0.4474 $\pm$ 0.0366} & \textcolor{gray}{304.9} & \textcolor{gray}{69.6}   \\
    jBOT-S (100\%)                   & 88.75 $\pm$ 0.40 & 0.9554 $\pm$ 0.0015 & 0.8712 $\pm$ 0.0097 & 0.4814 $\pm$ 0.0328 & 520.8 & 84.7   \\ \hline
    
    \textcolor{gray}{Sup.-B (10\%)}  & \textcolor{gray}{87.84 $\pm$ 0.51} & \textcolor{gray}{0.9467 $\pm$ 0.0023} & \textcolor{gray}{0.8429 $\pm$ 0.0132} & \textcolor{gray}{0.4131 $\pm$ 0.0285} & \textcolor{gray}{213.7} & \textcolor{gray}{56.1}  \\
    jBOT-B (10\%)                    & 88.62 $\pm$ 0.38 & 0.9542 $\pm$ 0.0017 & 0.8665 $\pm$ 0.0110 & 0.4843 $\pm$ 0.0306 & 409.8 & 85.3   \\

    \textcolor{gray}{Sup.-B (100\%)} & \textcolor{gray}{88.99 $\pm$ 0.35} & \textcolor{gray}{0.9569 $\pm$ 0.0018} & \textcolor{gray}{0.8756 $\pm$ 0.0072} & \textcolor{gray}{0.5021 $\pm$ 0.0331} & \textcolor{gray}{531.9} & \textcolor{gray}{96.9}  \\
    jBOT-B (100\%)                   & \textbf{89.11 $\pm$ 0.29} & \textbf{0.9584 $\pm$ 0.0015} & \textbf{0.8771 $\pm$ 0.0079} & \textbf{0.5122 $\pm$ 0.0389} & 531.9 & \textbf{109.6}   \\ \hline
\end{tabular}%
}
\end{table*}

\subsection{Downstream anomaly detection}
\label{sec:exp-anomaly}

For downstream anomaly detection, we use the embedding pre-trained only on QCD jets (q, g) as the normal data, and test on W, Z, and t jets as anomalous signals.
Although W, Z, and t jets are known physics signals rather than real anomalies, they are not exposed to the model at any point during training, so the setup is effectively signal-agnostic and suitable for anomaly detection evaluation.

We freeze the backbone encoder and map all examples in the training set to vectors in the \texttt{[CLS]} embedding, which form an in-distribution reference set.
We then define an anomaly score by computing a ``distance'' between each test example and the reference vectors in the embedding space, where an example far away from the reference bulk is more likely to be an anomalous signal.
We use four anomaly score metrics, namely $k$-NN, cosine similarity, Mahalanobis distance, and Gaussian mixture model (GMM).
Let $z(x)\in\mathbb{R}^{d_{\text{model}}}$ denote the \texttt{[CLS]} embedding of a jet $x$, and is $\ell_2$-normalized, and let $\mathcal{R}=\{z_{(i)}^{\text{ref}}\}_{i=1}^{M}$ be a reference set sampled from the training data.
The $k$-NN score is defined as the average Euclidean distance from the test jet $z(x)$ to its $k$ nearest neighbors $\{z_{(i|x)}^{\text{ref}}\}_{i=1}^{k}$ in $\mathcal{R}$:
\begin{equation}
    \label{eq:score-knn}
    s_{k\text{-NN}}(x;k)=\frac{1}{k}\sum_{i=1}^{k}\lVert z(x) - z_{(i|x)}^{\text{ref}} \rVert.
\end{equation}
Similarly, the cosine-similarity score measures angular distance to its $k$ nearest neighbors:
\begin{equation}
    s_{\text{cos}}(x;k)=-\tau\log\bigg(\frac{1}{k}\sum_{i=1}^{k}\exp\bigg(\frac{z(x)^{\text{T}}\cdot z_{(i|x)}^{\text{ref}}}{\tau}\bigg)\bigg),
\end{equation}
where $\tau=0.05$ is a temperature parameter.
The Mahalanobis distance~\cite{NEURIPS2018_abdeb6f5} fits class-conditional Gaussians with a tied covariance to the reference set and takes the minimum distance as the anomaly score:
\begin{equation}
    s_{\text{Maha}}(x)=\min_{c\in\{\text{q,g}\}}\big(z(x)-\mu_{c}\big)^{\text{T}}\Sigma^{-1}\big(z(x)-\mu_{c}\big),
\end{equation}
where $\mu_c$ is the class mean and $\Sigma$ is the shared covariance.
The GMM score fits a mixture of weighted Gaussians to the reference set and uses the negative log-likelihood as the anomaly score:
\begin{equation}
    s_{\text{GMM}}(x;K)=-\log\bigg(\sum_{i=1}^{K}w_{i}\mathcal{N}(z(x)|\mu_i,\Sigma_i)\bigg),
\end{equation}
where $K$ is the number of mixture components and $\{w_i, \mu_i, \Sigma_i\}_{i=1}^{K}$ are obtained by fitting the mixture model to the reference set.
We note that these metrics depend on the detailed clustering structure of the learned embedding, which can vary with randomness from the pre-training, so we pre-train ten jBOT-S models with identical configuration and report results from the best model.
We set $k=30$ for the $k$-NN and cosine similarity scores, and $K=4$ for the GMM score, as these values yield the highest performance on the combined signal for most models.

\begin{figure}[!t]
    \centering
    \includegraphics[width=0.4\textwidth]{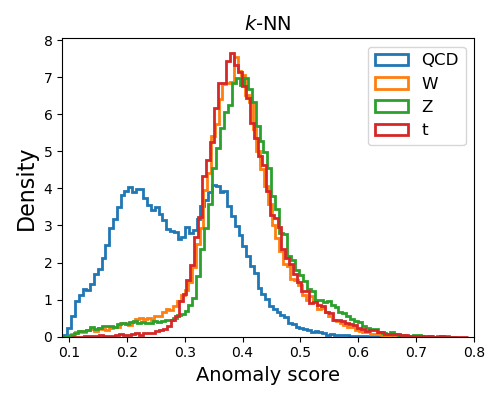}
    \includegraphics[width=0.4\textwidth]{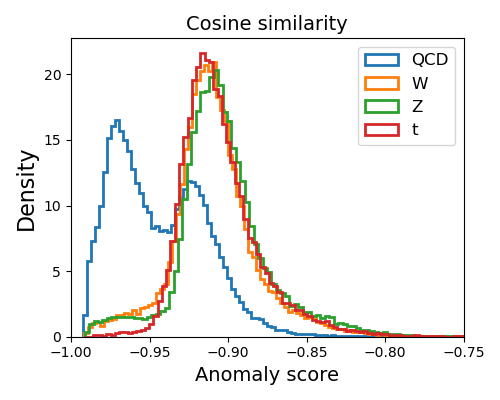} \\
    \includegraphics[width=0.4\textwidth]{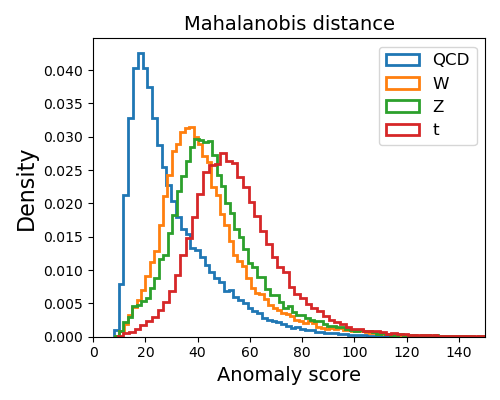}
    \includegraphics[width=0.4\textwidth]{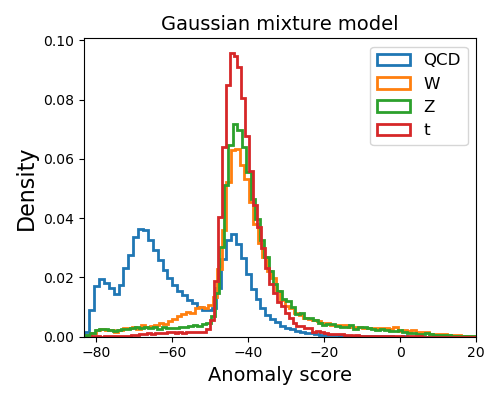}
    \caption{Anomaly score distributions: $k$-NN distance (upper left), cosine similarity (upper right), Mahalanobis distance (lower left), and GMM (lower right).}
    \label{fig:ad-score}
\end{figure}

\begin{figure}[!t]
    \centering
    \includegraphics[width=0.48\textwidth]{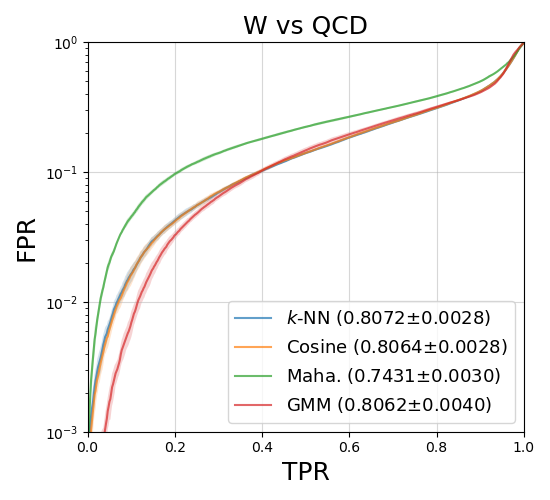}
    \includegraphics[width=0.48\textwidth]{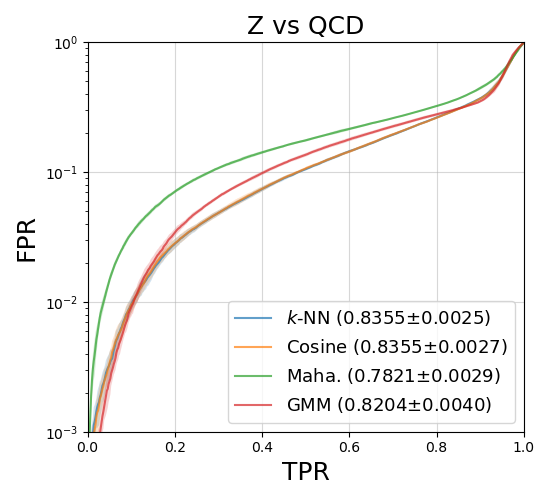} \\
    \includegraphics[width=0.48\textwidth]{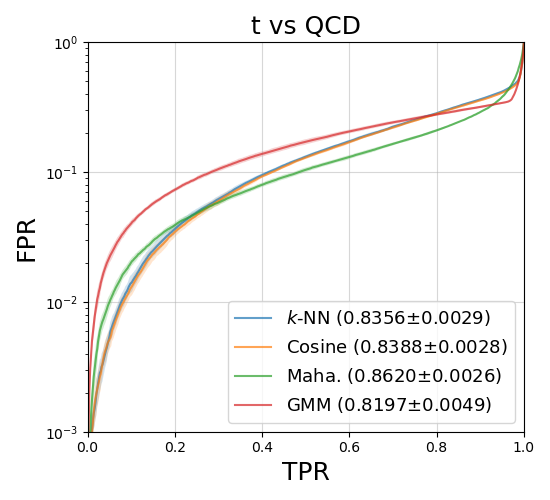}
    \includegraphics[width=0.48\textwidth]{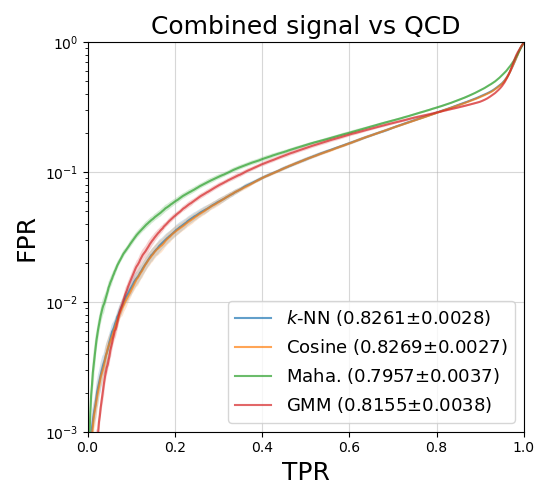}
    \caption{ROC curves for anomaly detection: W vs. QCD (upper left), Z vs. QCD (upper right), t vs. QCD (lower left), and the combined signal vs. QCD (lower right).}
    \label{fig:ad-roc}
\end{figure}

Fig.~\ref{fig:ad-score} shows the anomaly score distributions, where the QCD test sample consists of q and g jets in equal proportion.
The bimodal feature in the QCD distribution is a consequence of the separation between q and g clusters that emerges from pre-training (Fig.~\ref{fig:tsne}), which is resolved by score metrics that are sensitive to local structure in the embedding.
Fig.~\ref{fig:ad-roc} shows the ROC curves and Tab.~\ref{tab:anomaly} lists the AUCs for the individual signals and for the combined signal, where most anomalous signals are reasonably separated from the QCD background.
For a baseline comparison, we also quote results from Ref.~\cite{Hao:2022zns}, which uses reconstruction-based autoencoders for this task with different architectures, including a convolutional neural network (CNNAE), a graph neural network (GNNAE), and a Lorentz group equivariant network (LGAE).
Our method performance is broadly comparable to the reconstruction-based models and can perform better for some signals.
For example, our method using $k$-NN, cosine similarity, and GMM yields AUCs above 0.8 for the W and Z signals, while the reconstruction-based models yield AUCs below 0.8; for the t signal, our method using Mahalanobis distance yields the highest AUC among our metrics at around 0.86, compared to around 0.89 for CNNAE and GNNAE; for the combined signal, our method using $k$-NN, cosine similarity, and GMM yields AUCs of around 0.82, which is close to the highest AUC of 0.83 for LGAE.

It can also be seen that the performance varies with the choice of anomaly score.
This is expected because there is a high degree of freedom in defining the metric and hyperparameters for comparing a test example to a large reference set in a high-dimensional embedding space, and different choices probe different aspects of the structure: e.g., local structure for $k$-NN vs. global structure for GMM.
This flexibility introduces a large space to explore, and optimizations such as hyperparameter scans are required to select the best performing models.
Nonetheless, we show that measuring similarity between the test examples and the nominal examples in a self-supervised embedding pre-trained only on nominal data provides a viable anomaly detection strategy and can yield competitive or even better performance than common reconstruction-based methods.
These results serve as a proof-of-concept, and operationally useful working points on data would require scaling up the pre-training, dedicated augmentations, or combining the embedding with other scoring strategies.

\begin{table*}[!t]
\caption{Anomaly detection performance comparing jBOT (particle features only) and reconstruction-based autoencoder models from Ref.\cite{Hao:2022zns}.}
\label{tab:anomaly}
\centering
\resizebox{\textwidth}{!}{
\begin{tabular}{lcccc} \hline
    \multirow{2}{*}{Model} & \multicolumn{4}{c}{AUC}  \\
    & W & Z & t & Combined \\ \hline
    
    \textcolor{gray}{CNNAE}\cite{Hao:2022zns} & \textcolor{gray}{0.6886} & \textcolor{gray}{0.7247} & \textbf{0.8962} & \textcolor{gray}{0.7700} \\
    \textcolor{gray}{GNNAE}\cite{Hao:2022zns} & \textcolor{gray}{0.7558} & \textcolor{gray}{0.7805} & \textbf{0.8917} & \textcolor{gray}{0.8195} \\
    \textcolor{gray}{LGAE}\cite{Hao:2022zns}  & \textcolor{gray}{0.7489} & \textcolor{gray}{0.7909} & \textcolor{gray}{0.8669} & \textbf{0.8313} \\

    jBOT-S ($k$-NN) & \textbf{0.8072 $\pm$ 0.0028} & \textbf{0.8355 $\pm$ 0.0025} & 0.8356 $\pm$ 0.0029 & 0.8261 $\pm$ 0.0028 \\
    jBOT-S (Cosine) & \textbf{0.8064 $\pm$ 0.0028} & \textbf{0.8355 $\pm$ 0.0027} & 0.8388 $\pm$ 0.0028 & \textbf{0.8269 $\pm$ 0.0027} \\
    jBOT-S (Maha.)  & 0.7431 $\pm$ 0.0030 & 0.7821 $\pm$ 0.0029 & 0.8620 $\pm$ 0.0026 & 0.7957 $\pm$ 0.0037 \\
    jBOT-S (GMM)    & 0.8062 $\pm$ 0.0040 & 0.8204 $\pm$ 0.0040 & 0.8197 $\pm$ 0.0049 & 0.8155 $\pm$ 0.0038 \\ \hline

\end{tabular}%
}
\end{table*}

\section{Conclusion}
\label{sec:conclusion}

In this work, we have introduced jBOT, a self-supervised pre-training method based on the iBOT framework from computer vision, and applied it to jet data in HEP experiments at the CERN LHC.
We have shown that semantic clustering of different jet classes emerges in the jet representations learned via self-distillation objectives without supervision, and that probing the frozen embedding with $k$-NN or a linear classifier already achieves a five-class accuracy of around 70\%, compared to 76\% for a supervised model.
When the pre-trained model is fine-tuned on labeled data, it generally yields better performance than a supervised model trained from scratch, especially when the labeled dataset is small.
We have also shown that the frozen embedding from a self-supervised model trained only on unlabeled background jets can be used for anomaly detection, with the flexibility to define various metrics that can be competitive or better than common reconstruction-based autoencoder architectures.
We hope this work, as a new pre-training method based on self-distillation applicable to jets or similar physics data, contributes to ongoing developments in self-supervised learning for the HEP domain and potentially inspires adaptation within broader foundation model efforts.
Building on this proof-of-concept, future directions include scaling up pre-training on much larger unlabeled datasets, more robust anomaly detection strategies with dedicated augmentations, and fine-tuning on more diverse labeled data.

\section*{Acknowledgements}
DR is supported by the U.S. Department of Energy (DOE), Office of Science, Office of High Energy Physics Early Career Research program under Award No. DE-SC0025324.
This work used resources available through the National Research Platform (NRP) at the University of California, San Diego~\cite{10.1145/3708035.3736060}.
NRP has been developed, and is supported in part, by funding from National Science Foundation, from awards 1730158, 1540112, 1541349, 1826967, 2112167, 2100237, and 2120019, as well as additional funding from community partners.


\bibliography{references}

\end{document}